# Toward a Characterization of Uncertainty Measure for the Dempster-Shafer Theory

**David Harmanec**
Department of Systems Science and Industrial Engineering
Binghamton University – SUNY
Binghamton, New York 13902-6000
U.S.A.

## Abstract

This is a working paper summarizing results of an ongoing research project whose aim is to uniquely characterize the uncertainty measure for the Dempster-Shafer Theory. A set of intuitive axiomatic requirements is presented, some of their implications are shown, and the proof is given of the minimality of recently proposed measure $AU$ among all measures satisfying the proposed requirements.

## 1 Introduction.

Soon after the emergence of the Dempster-Shafer Theory (DST), researchers began their quest for a suitable measure of uncertainty (or information) for DST, which could be used in a similar manner as the Shannon entropy has been used within probability theory. However, the task is far from easy and has not been solved satisfactorily as yet. The common pitfall of the various proposed measures is the lack of property of subadditivity that is considered essential. (For more detailed discussion of the history of the search for uncertainty measure for DST see [7].)

In a recent paper [5], Harmanec and Klir proposed a new measure of uncertainty for DST. This measure is the only measure, among those introduced in the literature that satisfies all basic properties (including subadditivity) one would expect from such a measure. Independently, the measure was also proposed in [2] and [8], but with a somewhat different motivation.

Considering the fact that this measure is the only known measure of (total) uncertainty for DST that is additive, subadditive and in corresponding cases collapses into the Shannon and Hartley entropies, it is natural to ask whether these properties, possibly with some other intuitive properties, are sufficient to characterize this measure. An investigation of this problem is the topic of this paper. Unfortunately, I do not have a complete solution as yet, but I present some partial results. Namely, a set of eight plausible axiomatic requirements for a meaningful measure of uncertainty is suggested, some consequences of these requirements are shown, and the proof that the proposed measure is the smallest among all measures satisfying these axioms is given.

## 2 Notation and Basic Definitions.

Let $X$ denote some given finite (non-empty) universal set, usually called a *frame of discernment* in the context of DST. It is assumed that elements of $X$ represent all possible and mutually exclusive states of some system (answers to a question etc.). Let $\mathcal{P}(X)$ denote the power set of $X$.

A *belief function* is a function $Bel : \mathcal{P}(X) \longrightarrow [0,1]$ such that $Bel(\emptyset) = 0$, $Bel(X) = 1$, and

$$Bel(A_1 \cup \cdots \cup A_N) \geq$$
$$\sum \left\{ (-1)^{|I|+1} Bel(\cap_{i \in I} A_i) \,|\, \emptyset \neq I \subseteq \{1, \ldots, N\} \right\}$$

for all possible families of subsets of $X$. A belief function models a belief state of a Believer; $Bel(A)$ is the degree of belief of the Believer, on the basis of available evidence, that the actual state (the true answer, etc.) is in $A$.

A *basic probability assignment* is a function $m : \mathcal{P}(X) \longrightarrow [0,1]$ such that $m(\emptyset) = 0$ and

$$\sum_{A \in \mathcal{P}(X)} m(A) = 1.$$

A basic probability assignment is again considered as an evidence function; $m(A)$ is interpreted as a degree to which the Believer considers available evidence to support exactly $A$ and not any of its proper subsets. Any subset $A$ of $X$, for which $m(A) > 0$ is called a *focal element* of $m$.

As is well known [11], given a belief function, $Bel$, the function $m$ defined by the equation

$$m(A) = \sum_{B \subseteq A} (-1)^{|A-B|} Bel(B)$$

for all $A \in \mathcal{P}(X)$ is a basic probability assignment (called the basic probability assignment *associated* with $Bel$). Similarly, given a basic probability assignment $m$ the function $Bel$ defined by

$$Bel(A) = \sum_{B \subseteq A} m(B)$$

for all $A \in \mathcal{P}(X)$ is a belief function (called the belief function *associated* with $m$). Using this correspondence we can freely switch arguments about belief functions and basic probability assignments without worrying about inconsistency or misunderstanding.

Let $Y = \{E_1, E_2, \ldots, E_N\}$ denote a partition of $X$, i.e., $E_i \cap E_j = \emptyset$ for $i \neq j$ and $\cup_{i=1}^{N} E_i = X$. Then, for each $A \subseteq X$, the set $A \downarrow Y$ defined by

$$A \downarrow Y = \{E_i \in Y \mid E_i \cap A \neq \emptyset\}$$

is called *projection* of $A$ on $Y$. For a basic probability assignment $m$ on $X$, we define the *projection* of $m$ on $Y$ by the formula

$$m \downarrow Y(C) = \sum_{B \subseteq X \mid B \downarrow Y = C} m(B)$$

for all $C \in \mathcal{P}(Y)$. The reader can verify that $m \downarrow Y$ is really a basic probability assignment on $Y$; moreover,

$$Bel \downarrow Y(C) = Bel(\cup C),$$

for all $C \in \mathcal{P}(Y)$, is the corresponding belief function.

## 3 Axioms for Measure of Uncertainty

In this section, I discuss the properties a measure of uncertainty for DST should possess. I formulate the requirements in terms of basic probability assignments since it is simpler from a technical point of view, but belief functions or plausibility functions (see [11] for definition) could be used for this purpose as well.

**(R0) Functionality.** The sought uncertainty measure should be a mapping $\mathcal{U}$ that assigns a real number to each basic probability assignment $m$ on every finite frame of discernment $X$. For the purpose of this paper, I consider $m$ to be a set of ordered pairs $\langle A, m(A) \rangle$, where $A \in \mathcal{P}(X)$ and $m(A)$ is a basic probability number corresponding to $A$. Sometimes, I list only those pairs that correspond to the focal elements of $m$, and omit pairs with zero basic probability number. Formally,

$$\mathcal{U} : \mathbf{m} \longmapsto \mathcal{U}(\mathbf{m}),$$

for all $\mathbf{m} = \{\langle A, m(A) \rangle \mid A \in \mathcal{P}(X), m(\emptyset) = 0, \sum_{A \in \mathcal{P}(X)} m(A) = 1\}$ and all finite $X$'s, where $\mathcal{U}(\mathbf{m}) \in \mathbf{R}$.

**(R1) Label Independency.** The measure of uncertainty should not depend on the names of system's states (answers etc.). For example,

$$\mathbf{m}_1 = \{\langle \emptyset, 0 \rangle, \langle \{a\}, 0.2 \rangle, \langle \{b\}, 0.5 \rangle, \langle \{a, b\}, 0.3 \rangle\}$$

and

$$\mathbf{m}_2 = \{\langle \emptyset, 0 \rangle, \langle \{\triangle\}, 0.5 \rangle, \langle \{*\}, 0.2 \rangle, \langle \{\triangle, *\}, 0.3 \rangle\}$$

ought to have the same amount of uncertainty. Formally, let $X, Y$ be two finite sets such that $|X| = |Y|$ and let $\pi : X \longrightarrow Y$ be a one to one mapping of $X$ onto $Y$. Extend $\pi$ in the usual way onto power sets: $\pi(A) = \{\pi(x) \mid x \in A\}$ for all $A \in \mathcal{P}(X)$. Moreover, let $m$ be a basic probability assignment on $X$. Define a basic probability assignment $\pi(m)$ on $Y$ by

$$\pi(m)(B) = m(\pi^{-1}(B)),$$

for all $B \in \mathcal{P}(Y)$, where $\pi^{-1}$ denotes the inverse mapping for $\pi$. Then it is required that

$$\mathcal{U}(\mathbf{m}) = \mathcal{U}(\pi(\mathbf{m})).$$

Thanks to this requirement, we can restrict our considerations to one "canonical" frame of discernment with given cardinality $N$, e.g. $\{1, 2, \ldots, N\}$, without any loss of generality. Therefore, we can further simplify our notation. For any given natural number $N \geq 2$, we deal with a function of $2^N - 1$ variables

$$\mathcal{U}_N : \langle m_1, \ldots, m_N, m_{12}, \ldots, m_{1N}, \ldots, m_{12\ldots N} \rangle \longmapsto$$
$$\mathcal{U}_N(\langle m_1, \ldots, m_N, m_{12}, \ldots, m_{1N}, \ldots, m_{12\ldots N} \rangle) \in \mathbf{R},$$

which is defined on all $2^N - 1$-tuples such that $m_I \geq 0$, $I \subseteq \{1, 2, \ldots, N\}$ and $\sum_{\emptyset \neq I \subseteq \{1,2,\ldots,N\}} m_I = 1$. The empty set can be omitted since its basic probability number is always zero and, therefore, it does not have any influence on the value of $\mathcal{U}_N$. The requirement can now be reformulated as an requirement of symmetry.

**(R1') Symmetry.** For every $N$, every permutation $\pi$ of $\{1, 2, \ldots, N\}$, and every basic probability assignment $m$ on $\{1, 2, \ldots, N\}$.

$$\mathcal{U}_N(m_1, m_2, \ldots, m_{12\ldots N}) =$$
$$\mathcal{U}_N(m_{\pi(1)}, m_{\pi(2)}, \ldots, m_{\pi(1)\pi(2)\ldots\pi(N)}).$$

**(R2) Continuity.** Another property $\mathcal{U}$ should possess is continuity. This means, roughly, that a small change in basic probability assignment $m$ should not change abruptly the amount of uncertainty carried by $m$. More precisely, for any natural number $N \geq 2$, any basic probability assignment $m$ on $\{1, 2, \ldots, N\}$ and any $\emptyset \neq I, J \subseteq \{1, 2, \ldots, N\}$, the function

$$\mathcal{U}_N(m_1, \ldots, m_I - x, \ldots, m_J + x, \ldots, m_{12\ldots N})$$

as a function of $x$, defined for $x \in [0, m_I]$, is a continuous function on its domain.

**(R3) Expansibility.** This requirement states that adding an element to the frame of discernment without a change in Believer's beliefs (basic probability assignment) should not affect the amount of uncertainty contained in the beliefs. Formally,

$$\mathcal{U}_{N+1}(m_1, \ldots, m_N, 0, m_{12}, \ldots, m_{1N}, 0,$$
$$m_{23}, \ldots, m_{12\ldots N}, 0, \ldots, 0) =$$
$$\mathcal{U}_N(m_1, \ldots, m_N, m_{12}, \ldots, m_{1N}, m_{23}, \ldots, m_{12\ldots N})$$

for any basic probability assignment $m$ on $\{1, 2, \ldots, N\}$, where zeros in the left hand side of the equation are on the places corresponding to the subsets of $\{1, 2, \ldots, N+1\}$ containing $N+1$.

**(R4) Subadditivity.** Assume that the system described by the set of states $X$ can be divided into two subsystems with their respective sets of states being $Y_1$ and $Y_2$. Then by projecting our beliefs (basic probability assignment) onto $Y_1$ and $Y_2$ we preserve our knowledge (or information, if you wish) about subsystems, but we loose our knowledge about the interaction between the subsystems. Therefore, uncertainty contained in our knowledge of the whole system should not decrease by projecting it on subsystems.

Formally: Let $Y_1 = \{A_1, A_2, \ldots, A_P\}$ and $Y_2 = \{B_1, B_2, \ldots, B_Q\}$ be two distinct partitions of $X$, such that $|A_i \cap B_j| = 1$ for all $i \in \{1, 2, \ldots, P\}$ and all $j \in \{1, 2, \ldots, Q\}$. Then for an arbitrary basic probability assignment $m$ on $X$

$$\mathcal{U}(m) \leq \mathcal{U}(m \downarrow Y_1) + \mathcal{U}(m \downarrow Y_2).$$

**(R5) Additivity.** This requirement supplements the previous one. In the setting of the previous requirement, the overall uncertainty should be the same after the projections onto the subsystems if there is no interaction between subsystems.

Again, let $Y_1 = \{A_1, A_2, \ldots, A_P\}$ and $Y_2 = \{B_1, B_2, \ldots, B_Q\}$ be two (different) partitions of $X$, such that $|A_i \cap B_j| = 1$ for all $i \in \{1, 2, \ldots, P\}$ and all $j \in \{1, 2, \ldots, Q\}$. Moreover, let $m$ be a basic probability assignment on $X$ such that for every focal element $A$ of $m$,

$$A = \{y^1 \cap y^2 | y^1 \in A \downarrow Y_1, y^2 \in A \downarrow Y_2\}$$

and $m(A) = m \downarrow Y_1 (A \downarrow Y_1) \cdot m \downarrow (A \downarrow Y_2)$. Then,

$$\mathcal{U}(m) = \mathcal{U}(m \downarrow Y_1) + \mathcal{U}(m \downarrow Y_2).$$

**(R6) Monotone Dispensability.** According to this requirement, the uncertainty should not decrease after transferring part of basic probability mass from a focal element $A$ onto one of its supersets. In terms of belief functions, if we decrease our belief in a subset $A$ of $X$ without otherwise changing our beliefs (except those implied by our original belief in $A$) we should not decrease our uncertainty.

Formally, let $m$ be a basic probability assignment on $X$ and $A$ an arbitrary focal element of $m$. For any $B$, such that $A \subset B \subseteq X$, and any $\alpha \in [0, 1]$ define $m'$ as

$$m'(A) = \alpha \cdot m(A),$$
$$m'(B) = m(B) + (1 - \alpha) \cdot m(A),$$
$$m'(C) = m(C),$$

for all $C \in \mathcal{P}(X)$, $C \neq A$ and $C \neq B$. (Note that $m'$ is also a basic probability assignment.) Then

$$\mathcal{U}(m) \leq \mathcal{U}(m').$$

**(R7) Probabilistic Normalization.** To guarantee that the probabilistic (or conflict) component of uncertainty in DST [7] is measured in bits, it must hold that

$$\mathcal{U}_2\left(\frac{1}{2}, \frac{1}{2}, 0\right) = 1.$$

**(R8) Nonspecificity Normalization.** To guarantee that the nonspecificity component of uncertainty in DST [7] is measured in bits, it must hold that

$$\mathcal{U}_2(0, 0, 1) = 1.$$

## 4 Some Implications of the Requirements.

In this section, I derive some consequences of the eight requirements stated and discussed in Section 3.

**Theorem 1** *For any mapping $\mathcal{U}$ satisfying requirements (R1), (R3), and (R4) specified in Section 3 and any basic probability assignment $m$ on a finite frame of discernment $X$,*

$$\mathcal{U}(m) \geq 0.$$

**Proof.** The proof is a direct generalization of the corresponding proof in the characterization of the Shannon entropy [4]. It is enough to show that

$$\mathcal{U}_N(m_1, m_2, \ldots, m_N, m_{12}, \ldots, m_{12\ldots N}) \geq 0$$

for all $N \geq 2$ and all $(2^N - 1)$-tuples $\langle m_1, m_2, \ldots, m_N, m_{12}, \ldots, m_{12\ldots N} \rangle$ such that $m_I \geq 0$, $\emptyset \neq I \subseteq \{1, 2, \ldots, N\}$ and $\sum_{\emptyset \neq I \subseteq \{1, 2, \ldots, N\}} m_I = 1$.

By a repeated use of the expansibility requirement (R3), we have

$$\mathcal{U}_N(m_1, m_2, \ldots, m_N, m_{12}, \ldots, m_{12\ldots N}) = \\ \mathcal{U}_{N+1}(m_1, m_2, \ldots, m_N, 0, \\ m_{12}, \ldots, m_{1N}, 0, \ldots, m_{12\ldots N}, 0, \ldots, 0) = \\ \vdots \\ = \mathcal{U}_{N^2}(m_1, m_2, \ldots, m_N, 0, \ldots, 0, \\ m_{12}, \ldots, m_{1N}, 0, \ldots, 0, \ldots, m_{12\ldots N}, 0, \ldots, 0), \quad (1)$$

and by subadditivity (R4) and symmetry (R1), we get

$$\mathcal{U}_{N^2}(m_1, m_2, \ldots, m_N, 0, \ldots, 0, \\ m_{12}, \ldots, m_{1N}, 0, \ldots, 0, \ldots, m_{12\ldots N}, 0, \ldots, 0) \leq \\ \mathcal{U}_N(m_1, m_2, \ldots, m_N, m_{12}, \ldots, m_{12\ldots N}) + \\ \mathcal{U}_N(m_1, m_2, \ldots, m_N, m_{12}, \ldots, m_{12\ldots N}). \quad (2)$$

(To better see this step, one can imagine instead of $\{1,2,\ldots,N^2\}$ the Cartesian product $\{1,2,\ldots,N\} \times \{1,2,\ldots,N\}$; the (potentially) non-zero elements $m_1, \ldots, m_{12\ldots N}$ then correspond to the "diagonal" elements of the Cartesian product, as illustrated in Figure 1.) From (1) and (2), we may conclude

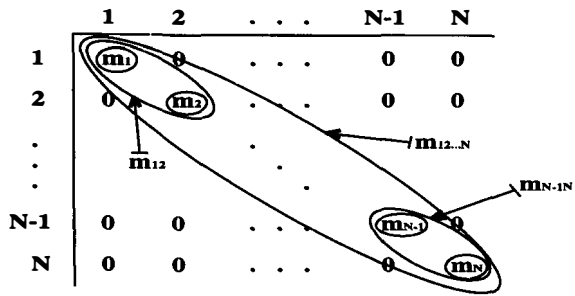

Figure 1: Imagine $\{1,2,\ldots,N\} \times \{1,2,\ldots,N\}$ instead of $\{1,2,\ldots,N^2\}$

$$\mathcal{U}_N(m_1, m_2, \ldots, m_N, m_{12}, \ldots, m_{12\ldots N}) \geq 0. \quad \blacksquare$$

Next I show that $\mathcal{U}$ collapses to the Shannon entropy and to the Hartley entropy in relevant situations.

**Theorem 2** *Assume that $N \geq 2$, $m_i \geq 0$ for $i \in \{1,2,\ldots,N\}$ and $\sum_{i=1}^N m_i = 1$. Then, for any mapping $\mathcal{U}$ satisfying requirements (R1)–(R5), and (R7) specified in Section 3, it holds that*

$$\mathcal{U}_N(m_1, m_2, \ldots, m_N, 0, \ldots, 0) = -\sum_{i=1}^N m_i \log_2 m_i.$$

**Proof.** Forte [4] showed that the only function $\mathcal{K}_N$ defined on $N$-tuples $\langle p_1, p_2, \ldots, p_N \rangle$ such that $p_i \geq 0$, $i \in \{1, 2, \ldots, N\}$ and $\sum_{i=1}^N p_i = 1$, which is symmetric, expansible, additive, subadditive, satisfying the normalization $\mathcal{K}_2\left(\frac{1}{2}, \frac{1}{2}\right) = 1$ and the condition

$$\lim_{p \to 0+} \mathcal{K}_2(p, 1-p) = \mathcal{K}_2(0,1), \quad (3)$$

is the Shannon entropy. Since our requirements of symmetry, expansibility, additivity, subadditivity and probabilistic normalization corresponds under the assumptions of the theorem exactly to the respective Forte's axioms, and since the continuity requirement implies the condition (3), the theorem follows. $\blacksquare$

**Theorem 3** *For any mapping $\mathcal{U}$ satisfying requirements (R1), (R3), (R5), (R6), and (R8)*

$$\mathcal{U}(\langle A, 1 \rangle) = \log_2 |A|,$$

*for arbitrary finite frame of discernment $X$ and arbitrary $A \subseteq X$.*

**Proof.** Rényi [10] showed that the only function $\mathcal{I}$ defined on natural numbers ($\geq 1$), which satisfies
(a) $\mathcal{I}(N.M) = \mathcal{I}(N) + \mathcal{I}(M)$, for all $N, M \in \mathbf{N}$,
(b) $\mathcal{I}(N) \leq \mathcal{I}(N+1)$, for all $N \in \mathbf{N}$,
(c) $\mathcal{I}(2) = 1$,
is the Hartley entropy. However, from the label independency and the expansibility we know that $\mathcal{U}(\langle A, 1 \rangle)$ depends only on the cardinality of $A$, so we are looking for a function $f$ defined on natural numbers ($\geq 1$), such that

$$\mathcal{U}(\langle A, 1 \rangle) = f(|A|).$$

From additivity we know that $f$ satisfies (a), from monotone dispensability it has to obey (b), and non-specificity normalization guarantees (c). Therefore the theorem follows from Rényi's result. $\blacksquare$

**Corollary 4** *For any mapping $\mathcal{U}$ satisfying requirements (R1), (R3), (R5), (R6), and (R8), and any basic probability assignment $m$ on a finite frame of discernment $X$,*

$$\mathcal{U}(m) \leq \log_2 |X|.$$

**Proof.** It is enough to consider $X = \{1, 2, \ldots, N\}$. By repeated use of monotone dispensability, we get

$$\mathcal{U}_N(m) \leq \\ \mathcal{U}_N(0, m_2, m_3, \ldots, m_N, m_{12}, \ldots, m_{12\ldots N} + m_1) \leq \\ \vdots \\ \leq \mathcal{U}_N(0, 0, \ldots, 1).$$

Then, by Theorem 3,

$$\mathcal{U}_N(0, 0, \ldots, 1) = \log_2 N. \quad \blacksquare$$

## 5 Minimality of the Uncertainty Measure $AU$

Before proving that the proposed measure $AU$ is minimal, let me show that $AU$ satisfies the requirements (R0) - (R8) and, therefore, at least one uncertainty measure satisfies all these requirements.

**Definition 1** *[5] Let $X$ denote a (finite) frame of discernment and Bel a belief function defined on $X$. We define the measure of uncertainty contained in Bel, denoted as $AU(Bel)$, by*

$$AU(Bel) = \max\left\{-\sum_{x \in X} p_x \log_2 p_x\right\},$$

*where the maximum is taken over all $\{p_x\}_{x \in X}$ such that $p_x \in [0,1]$ for all $x \in X$, $\sum_{x \in X} p_x = 1$, and for all $A \subseteq X$, $Bel(A) \leq \sum_{x \in A} p_x$.*

Note that the maximum always exists since we are maximizing a continuous function on a simplex[1]. To be able to take an advantage of the basic probability assignment, we need the following theorem.

**Theorem 5** *(Dempster [3]) Let $X$ be a frame of discernment, Bel a belief function on $X$ and $m$ the corresponding basic probability assignment; then a tuple $\langle p_x \rangle_{x \in X}$ satisfies the following constraints*

$$0 \leq p_x \leq 1 \quad \text{for all } x \in X,$$
$$\sum_{x \in X} p_x = 1, \text{ and}$$
$$Bel(A) \leq \sum_{x \in A} p_x, \text{ for all } A \subseteq X,$$

*if and only if there exist non-negative real numbers $\alpha_A^x$ for all $\emptyset \neq A \subseteq X$ and all $x \in A$ such that*

$$p_x = \sum_{A | x \in A} \alpha_A^x$$

*and*

$$\sum_{x | x \in A} \alpha_A^x = m(A).$$

**Theorem 6** *Under the conditions of Definition 1, the measure $AU$ satisfies all the requirements (R0) - (R8).*

**Proof.** Requirements (R1) - (R3) follow directly from the definition of $AU$. In our previous paper [5], we showed that $AU$ satisfies requirements (R4) and (R5). In the same paper we have also shown that $AU$ collapses to the Shannon entropy and the Hartley entropy in the corresponding cases. This fact implies requirements (R7) and (R8). It only remains to show that $AU$ obeys the requirement (R6).

[1] We assume $0 \log_2 0 = 0$.

Let $X$, $A$, $B$, $\alpha$, $m$, and $m'$ have the same meaning as in the assumptions of the requirement (R6). Let $\langle p_x \rangle_{x \in X}$ be such a tuple that $p_x \geq 0$, for $x \in X$, $\sum_{x \in X} p_x = 1$, for all $C \subseteq X$ $\sum_{x \in C} p_x \geq Bel(C)$, where Bel denotes the belief function corresponding to the basic probability assignment $m$, and $-\sum_{x \in X} p_x \log_2 p_x = AU(m)$. Then by Theorem 5, there exist non-negative real numbers $\beta_C^x$ for all $\emptyset \neq C \subseteq X$ and all $x \in C$ such that

$$p_x = \sum_{C | x \in C} \beta_C^x$$

and

$$\sum_{x | x \in C} \beta_C^x = m(C).$$

Put $\beta_A'^x = \alpha . \beta_A^x$ for all $x \in A$, $\beta_B'^x = \beta_B^x + (1-\alpha) . \beta_A^x$ for all $x \in A$, $\beta_B'^x = \beta_B^x$ for all $x \in B - A$, and $\beta_C'^x = \beta_C^x$ for all $\emptyset \neq C \subseteq X$, $C \neq A$, $C \neq B$, and all $x \in C$. Since

$$\sum_{C | x \in C} \beta_C'^x = \sum_{C | x \in C} \beta_C^x = p_x$$

if $x \notin A$, and

$$\sum_{C | x \in C} \beta_C'^x = \beta_A'^x + \beta_B'^x + \sum_{C | x \in C, C \neq A, C \neq B} \beta_C'^x =$$
$$\alpha . \beta_A^x + \beta_B^x + (1 - \alpha) . \beta_A^x + \sum_{C | x \in C, C \neq A, C \neq B} \beta_C^x =$$
$$\sum_{C | x \in C} \beta_C^x = p_x$$

if $x \in A$, we have

$$\sum_{x | x \in A} \beta_A'^x =$$
$$\sum_{x | x \in A} \alpha . \beta_A^x =$$
$$\alpha . \sum_{x | x \in A} \beta_A^x =$$
$$\alpha . m(A) = m'(A),$$
$$\sum_{x | x \in B} \beta_B'^x =$$
$$\sum_{x | x \in B-A} \beta_B^x + \sum_{x | x \in A} (\beta_B^x + (1-\alpha) . \beta_A^x) =$$
$$\sum_{x | x \in B} \beta_B^x + (1-\alpha) . \sum_{x | x \in A} \beta_A^x =$$
$$m(B) + (1-\alpha) . m(A) = m'(B),$$

and

$$\sum_{x | x \in C} \beta_C'^x = \sum_{x | x \in C} \beta_C^x = m(C) = m'(C),$$

for all $\emptyset \neq C \subseteq X$, $C \neq A$, $C \neq B$, we get by Theorem 5 and definition of $AU$

$$-\sum_{x \in X} p_x \log_2 p_x \leq AU(m'),$$

which means

$$AU(m) \leq AU(m').$$

This concludes the proof of the theorem. ∎

Now we know that there is at least one uncertainty measure satisfying all the requirements (R0) - (R8).

The question is, whether it is also the only one with this property. I do not know the answer to this question as yet, but I can show that $AU$ is the smallest measure satisfying all the requirements.

**Theorem 7** *Let $X$ denote a (finite) frame of discernment and $m$ a basic probability assignment on $X$. Then for any mapping $\mathcal{U}$ satisfying all the requirements (R0) - (R8), we have*

$$AU(m) \leq \mathcal{U}(m).$$

**Proof.** Let $Bel$ denote the belief function on $X$ corresponding to $m$. Consider arbitrary $\langle p_x \rangle_{x \in X}$, $p_x \in [0,1]$, $\sum_{x \in X} p_x = 1$, such that

$$Bel(A) \leq \sum_{x \in A} p_x,$$

for all $A \subseteq X$. By Theorem 2

$$\begin{aligned}AU(\{\langle \{x\}, p_x\rangle \mid x \in X\}) = \\ -\sum_{x \in X} p_x \log_2 p_x = \\ \mathcal{U}(\{\langle \{x\}, p_x\rangle \mid x \in X\}).\end{aligned} \quad (4)$$

From Theorem 5, we know that there are non-negative real numbers $\alpha_A^x$ for all $\emptyset \neq A \subseteq X$ and all $x \in A$ such that

$$p_x = \sum_{A \mid x \in A} \alpha_A^x,$$

for all $x \in X$, and

$$\sum_{x \mid x \in A} \alpha_A^x = m(A),$$

for all $\emptyset \neq A \subseteq X$. Let $X = \{x_1, x_2, \ldots, x_N\}$. Then by repeated use of the monotone dispensability we get

$$\mathcal{U}(\{\langle \{x_i\}, p_{x_i}\rangle \mid i \in \{1,2,\ldots,N\}\}) = $$
$$\mathcal{U}\left(\left\{\left\langle \{x_i\}, \sum_{A \mid x_i \in A} \alpha_A^{x_i}\right\rangle \mid i \in \{1,2,\ldots,N\}\right\}\right) \leq $$
$$\mathcal{U}(\{\left\langle\{x_1\}, p_{x_1} - \alpha_{\{x_1,x_2\}}^{x_1}\right\rangle, \left\langle\{x_1,x_2\}, \alpha_{\{x_1,x_2\}}^{x_1}\right\rangle\} $$
$$\cup \{\langle\{x_i\}, p_{x_i}\rangle \mid i \in \{2,3,\ldots,N\}\}) \leq $$
$$\vdots$$
$$\leq \mathcal{U}(\{\left\langle\{x_1\}, \alpha_{\{x_1\}}^{x_1}\right\rangle, \left\langle\{x_1,x_2\}, \alpha_{\{x_1,x_2\}}^{x_1}\right\rangle, \ldots, $$
$$\langle X, \alpha_X^{x_1}\rangle\} \cup \{\langle\{x_i\}, p_{x_i}\rangle \mid i \in \{2,3,\ldots,N\}\}) \leq $$
$$\vdots$$
$$\leq \mathcal{U}\left(\left\{\left\langle A, \sum_{x \mid x \in A}\alpha_A^x\right\rangle \mid \emptyset \neq A \subseteq X\right\}\right) = $$
$$\mathcal{U}(\{\langle A, m(A)\rangle \mid \emptyset \neq A \subseteq X\}) = \mathcal{U}(m) \quad (5)$$

It follows from (4) and (5) that

$$AU(\{\langle\{x\}, p_x\rangle \mid x \in X\}) \leq \mathcal{U}(m) \quad (6)$$

for any $\langle p_x\rangle_{x \in X}$, $p_x \in [0,1]$, $\sum_{x \in X} p_x = 1$, such that $Bel(A) \leq \sum_{x \in A} p_x$. However, the inequality (6) still holds if we take maximum over all acceptable $\langle p_x\rangle_{x \in X}$ on the left hand side

$$\max AU(\{\langle\{x\}, p_x\rangle \mid x \in X\}) \leq \mathcal{U}(m).$$

We may conclude the proof of the theorem by realizing that

$$\max AU(\{\langle\{x\}, p_x\rangle \mid x \in X\}) = AU(m). \quad \blacksquare$$

## 6 Conclusion

I presented and tried to justify a set of eight requirements, which I deem essential for any reasonable measure of uncertainty for DST. I proved some consequences of these requirements and showed that the recently proposed measure $AU$ [5] satisfies all the required axioms. The question, whether $AU$ is the only measure satisfying these requirements or not remains an open problem. Depending on the solution of this problem, we can look at Theorem 7 from two different angles. It can be either considered as a step toward uniqueness proof or it can be considered as a justification for choosing $AU$ as *the* measure of uncertainty for DST. It seems reasonable to take the "smallest" measure, since we do not want to attach to a belief function more uncertainty than necessary. Though this requirement is probably less appealing than the requirements (R0) - (R8), it looks like a good practical guidance in the choice of the uncertainty measure for DST.

**Acknowledgment**

I am grateful to Professor G. Klir and two anonymous referees for their comments on the manuscript of this paper.

The work on this paper was supported by ONR grant No. N00014-94-1-0263.

## References

[1] J. Aczel, B. Forte, and C. T. Ng. Why the Shannon and Hartley entropies are 'natural'. *Advances in Applied Probability*, 6(1):147–158, 1974.

[2] C. W. R. Chau, P. Lingras, and S. K. M. Wong. Upper and lower entropies of belief functions using compatible probability functions. In J. Komorowski and Z. W. Raś, editors, *Methodologies for Intelligent Systems*, number 689 in LNAI, pages 306–315. Springer-Verlag, 1993. Proceedings of 7th International Symposium ISMIS'93.

[3] A. P. Dempster. Upper and lower probabilities induced by a multi-valued mapping. *Annals of Mathematical Statistics*, 38(2):325–339, 1967.

[4] B. Forte. Why Shannon's entropy. In *Symposia Mathematica*, volume XV, pages 137–152. Academic Press, New York, Instituto Nazionale di Alta Mathematica edition, 1975.

[5] D. Harmanec and G. J. Klir. Measuring total uncertainty in Dempster-Shafer theory: A novel approach. *International Journal of General Systems*, 22(4):405–419, 1994.

[6] R. V. Hartley. Transmission of information. *Bell System Technical Journal*, 7:535–563, 1928.

[7] G. J. Klir. Developments in uncertainty - based information. In M. C. Yovits, editor, *Advances in Computers*, volume 36, pages 255–332. Academic Press, San Diego, 1993.

[8] Y. Maeda and H. Ichihashi. An uncertainty measure with monotonicity under the random set inclusion. *International Journal of General Systems*, 21(4):379–392, 1993.

[9] Y. Maeda, H. T. Nguyen, and H. Ichihashi. Maximum entropy algorithms for uncertainty measures. *International Journal of Uncertainty, Fuzziness and Knowledge-Based Systems*, 1(1):69–93, 1993.

[10] A. Rényi. *Probability Theory*. North-Holland, Amsterdam, 1970.

[11] G. Shafer. *A Mathematical Theory of Evidence*. Princeton University Press, Princeton, 1976.

[12] C. E. Shannon. A mathematical theory of communication. *Bell Technical Journal*, 4:379–423, 1948.